# Eve: A Gradient Based Optimization Method with Locally and Globally Adaptive Learning Rates

Hiroaki Hayashi[1,*]   Jayanth Koushik[1,*]   Graham Neubig[1]

## Abstract

Adaptive gradient methods for stochastic optimization adjust the learning rate for each parameter locally. However, there is also a global learning rate which must be tuned in order to get the best performance. In this paper, we present a new algorithm that adapts the learning rate locally for each parameter separately, and also globally for all parameters together. Specifically, we modify Adam, a popular method for training deep learning models, with a coefficient that captures properties of the objective function. Empirically, we show that our method, which we call *Eve*, outperforms Adam and other popular methods in training deep neural networks, like convolutional neural networks for image classification, and recurrent neural networks for language tasks.

## 1 Introduction

Training deep neural networks is a challenging non-convex optimization problem. Stochastic Gradient Descent (SGD) is a simple way to move towards local optima by following the negative (sub)gradient. However, vanilla SGD is slow in achieving convergence for large-scale problems. One issue arises from the use of a global learning rate, which is difficult to set. To prevent the loss from "bouncing around" or diverging in directions with high curvature, the learning rate must be kept small. But this leads to slow progress in directions with low curvature. In many problems, the sparsity of gradients creates an additional challenge for SGD. Some parameters might be used very infrequently, but still be very informative. Therefore, these parameters should be given a large learning rate when observed.

The issues highlighted above motivate the need for adaptive learning rates that are local to each parameter in the optimization problem. While there has been much work in this area, here we focus on the family of methods based on the Adagrad algorithm[2]. These methods maintain a separate learning rate for each parameter; and these local learning rates are made adaptive using some variation of the sum of squared gradients. Roughly speaking, if the gradients for some parameter have been large, its learning rate is reduced; and if the gradients have been small, the learning rate is increased. Variations of Adagrad such as RMSprop[15], Adadelta[17], or Adam[6] are, by far, the most popular alternatives to vanilla SGD for training deep neural networks.

In addition to parameter-specific local learning rates, the adaptive methods described above also have a global learning rate which determines the overall step size. In many cases, this global learning rate is left at its default value; however, to get the best performance, it needs to be tuned, and also adapted throughout the training process. A common strategy is to decay the learning rate over time, which adds an additional hyperparameter, the decay strength, that needs to be chosen carefully.

In this work, we address the problem of adapting the global learning rate with a simple method that incorporates "feedback" from the objective function. Our algorithm, Eve, introduces a scalar coefficient $d_t$ which is used to adapt the global learning rate to be $\alpha_t = \alpha_1/d_t$, where $\alpha_1$ is the initial learning rate. $d_t$ depends on the history of stochastic objective function evaluations, and captures two properties of its behavior: variation in consecutive iterations and sub-optimality. Intuitively, high variation should reduce the learning rate and high sub-optimality should increase the learning rate. We specifically apply this idea to Adam[6], a popular method for training deep neural networks.

## 2 Related work

Our work builds on recent advancements in gradient based optimization methods with locally adaptive learning rates. Notable members in this family are Adagrad[2], Adadelta[17], RMSProp[15], Adam (and Adamax)[6]. These methods adapt the learning rate using sum of squared gradients, an estimate of the uncentered second moment. Some of these methods also use momentum, or running averages instead of the raw gradient. Being first-order methods, they are simple to implement, and computationally efficient. In practice, they perform very well and are the methods of choice for training large neural networks.

As was discussed in the introduction, these methods have a global learning rate which is generally constant or annealed over time. Two popular decay algorithms are exponential decay which sets $\alpha_t = \alpha_1 \exp(-\gamma t)$, and $1/t$ decay which sets $\alpha_t = \alpha_1/(1 + \gamma t)$. Here $\gamma$ is the decay strength, $t$ is iteration number, and $\alpha_1$ is the initial learning rate. For Adam, Kingma and Ba (2014)[6] suggest $1/\sqrt{t}$ decay which sets $\alpha_t = \alpha_1/\sqrt{1+\gamma t}$. We compare our proposed method with Adam using these decay algorithms. There are other heuristic scheduling algorithms used in practice like reducing the learning rate by some factor after every some number of iterations, or reducing the learning rate when the loss on a held out validation set stalls. Smith (2017)[13] proposes a schedule where the learning rate is varied cyclically between a range of values.

---

[1]Carnegie Mellon University. *Equal contribution. Email: {hiroakih,jkoushik,gneubig}@cs.cmu.edu.

Schaul et al. (2013)[12] take up the more ambitious goal of completely eliminating the learning rate parameter, with a second-order method that computes the diagonal Hessian using the bbprop algorithm[8]. Note, however, that the method is only applicable under mean squared error loss. Finally, for convex optimization, Polyak (1987)[11] proposes a way to select step sizes for the subgradient method when the optimum is known. It can be shown that with steps of size $(f_t - f^\star)/\|g_t\|^2$ (where $g_t$ is a subgradient), the subgradient method converges to the optimal value under some conditions. Our method also makes use of the optimal value $f^\star$ for adapting the global learning rate.

## 3 Method

### 3.1 PRELIMINARIES: ADAM

Since our method builds on top of the Adam algorithm, we begin with a brief description of the method. First, we establish the notation: let $f(\theta)$ be a stochastic objective function with parameters $\theta$, and let $\theta_t$ be the value of the parameters after time $t$. Let $f_t = f(\theta_t)$ and $g_t = \nabla_\theta f(\theta_t)$. Adam maintains a running average of the gradient given by

$$m_t = \beta_1 m_{t-1} + (1-\beta_1)g_t, \qquad (1)$$

with $m_0 = 0$. A correction term is applied to remove the bias caused by initializing with 0.

$$\widehat{m}_t = m_t/(1-\beta_1^t). \qquad (2)$$

$\widehat{m}_t$ is an unbiased estimate of the gradient's first moment assuming stationarity ($\mathbb{E}[\widehat{m}_t] = \mathbb{E}[g_t]$). A similar term is computed using the squared gradients:

$$\begin{aligned} v_t &= \beta_2 v_{t-1} + (1-\beta_2)g_t^2. \\ \widehat{v}_t &= v_t/(1-\beta_2^t). \end{aligned} \qquad (3)$$

$\widehat{v}_t$ is an unbiased estimate of the gradient's uncentered second moment ($\mathbb{E}[\widehat{v}_t] = \mathbb{E}[g_t^2]$). Then, Adam updates parameters using the update equation

$$\theta_{t+1} = \theta_t - \alpha_t \frac{\widehat{m}_t}{\sqrt{\widehat{v}_t} + \epsilon}. \qquad (4)$$

### 3.2 PROPOSED METHOD: EVE

Our method is motivated by simple intuitive arguments. Let $f$ denote the stochastic objective function that needs to be minimized. Let $f_t$ be it's value at iteration $t$, and let $f^\star$ be its global minimum. First, consider the quantity $|f_t - f_{t-1}|$. This captures the variability in the objective function i.e., how much the function is changing from one step to the other. If this quantity is large, any particular stochastic evaluation of $f$ should be given less "weight", and the step taken based on it should be small. So we would like the learning rate to depend inversely on $|f_t - f_{t-1}|$. Next, consider the quantity $f_t - f^\star$, where $f^\star$ is the expected global minimum of $f$. This is the sub-optimality i.e., it denotes how far we are from the minimum at time $t$. Intuitively, far from the minimum, we should take big steps to make quick progress, and if we are close to the minimum, we should take small steps. Hence we would like the learning rate to depend directly on $f_t - f^\star$. Putting these two together, our method scales the learning rate by $(f_t - f^\star)/|f_t - f_{t-1}|$:

$$\alpha_t = \frac{\alpha_1}{d_t} = \alpha_1 \frac{f_t - f^\star}{|f_t - f_{t-1}|}. \qquad (5)$$

However, this simple method is not stable because once the loss increases, the increase of numerator in Equation 5 directly increases the learning rate. This can result in the learning rate blowing up due to taking an even bigger step, causing the numerator to further increase, and so forth. To prevent this, we make modifications to stabilize the update rule by first replacing $f_t$ in the numerator of Equation 5 with $\min\{f_t, f_{t-1}\}$. This will reduce the chance of the vicious cycle mentioned above by keeping the numerator at the same value if the loss increases. In addition, we clip the term with a range parameterized by a constant $c$ to avoid extreme values.

$$\widehat{d}_t = \text{clip}(d_t, [1/c, c]). \qquad (6)$$

Finally, for smoothness, we take a running average of the clipped $d_t$

$$\widetilde{d}_t = \beta_3 \widetilde{d}_{t-1} + (1-\beta_3)\widehat{d}_t. \qquad (7)$$

The learning rate is then given by $\alpha_t = \alpha_1/\widetilde{d}_t$. Thus, the learning rate will be in the range $[\alpha_1/c, c\alpha_1]$. Combining this with the Adam update rule gives our complete algorithm, which we call Eve, and is shown in Figure 1. Below, the equations for computing $\widetilde{d}_t$ are summarized. We set

```
 1: α₁ = 10⁻³                              ▷ default
 2: β₁ = 0.9, β₂ = 0.999, β₃ = 0.999       ▷ default
 3: c = 10                                  ▷ default
 4: ε = 10⁻⁸                                ▷ default

 5: m₀ = v₀ ← 0
 6: t ← 0

 7: while stopping condition is not reached do
 8:     t ← t + 1
 9:     gₜ ← ∇_θ f(θₜ)
10:     mₜ ← β₁mₜ₋₁ + (1 − β₁)gₜ
11:     m̂ₜ ← mₜ/(1 − β₁ᵗ)
12:     vₜ ← β₂vₜ₋₁ + (1 − β₂)gₜ²
13:     v̂ₜ ← vₜ/(1 − β₂ᵗ)

14:     if t > 1 then
15:         dₜ ← |fₜ − fₜ₋₁| / (min{fₜ, fₜ₋₁} − f★)
16:         d̂ₜ ← clip(dₜ, [1/c, c])
17:         d̃ₜ ← β₃d̃ₜ₋₁ + (1 − β₃)d̂ₜ
18:     else
19:         d̃ₜ ← 1
20:     end if

21:     θₜ ← θₜ₋₁ − (α₁/d̃ₜ) · m̂ₜ/(√v̂ₜ + ε)
22: end while
23: return θₜ
```

Figure 1: Eve algorithm. Wherever applicable, products are element-wise.

$\widetilde{d}_1 = 1$, and for $t > 1$, we have

$$\begin{aligned}
d_t &= \frac{|f_t - f_{t-1}|}{\min\{f_t, f_{t-1}\} - f^\star}. \\
\widehat{d}_t &= \text{clip}(d_t, [1/c, c]). \\
\widetilde{d}_t &= \beta_3 \widetilde{d}_{t-1} + (1 - \beta_3)\widehat{d}_t.
\end{aligned} \qquad (8)$$

### 3.3 DISCUSSION OF LIMITATIONS

One condition of our method, which can also be construed as a limitation, is that it requires knowledge of the global minimum of the objective function $f^\star$. However, the method still remains applicable to a large class of problems. Particularly in deep learning, regularization is now commonly performed indirectly with dropout[14] or batch normalization[5] rather than weight decay. Therefore, under mean squared error or cross-entropy loss functions, the global minimum is simply 0. This will be the case for all our experiments, and we show that Eve can improve over other methods in optimizing complex, practical models.

## 4  Experiments

Now we conduct experiments to compare Eve with other popular optimizers used in deep learning. We use the same hyperparameter settings (as described in Figure 1) for all experiments. We also conduct an experiment to study the behavior of Eve with respect to the new hyperparameters $\beta_3$ and $c$. For each experiment, we use the same random number seed when comparing different methods. This ensures same weight initializations (we use the scheme proposed by Glorot and Bengio (2010)[3] for all experiments), and mini-batch splits. In all experiments, we use cross-entropy as the loss function, and since the models don't have explicit regularization, $f^\star$ is set to 0 for training Eve.

### 4.1 TRAINING CNNS

First we compare Eve with other optimizers for training a Convolutional Neural Network (CNN). The optimizers we compare against are Adam, Adamax, RMSprop, Adagrad, Adadelta, and SGD with Nesterov momentum[10] (momentum 0.9). The learning rate was searched over $\{1 \times 10^{-6}, 5 \times 10^{-6}, 1 \times 10^{-5}, 5 \times 10^{-5}, 1 \times 10^{-4}, 5 \times 10^{-4}, 1 \times 10^{-3}, 5 \times 10^{-3}, 1 \times 10^{-2}, 5 \times 10^{-2}, 1 \times 10^{-1}\}$, and the value which led to the lowest final loss was selected for reporting results. For Adagrad, Adamax, and Adadelta, we additionally searched over the prescribed default learning rates ($10^{-2}$, $2 \times 10^{-3}$, and 1 respectively).

The model is a deep residual network[4] with 16 convolutional layers. The network is regularized with batch normalization and dropout, and contains about 680,000 parameters, making it representative of a practical model.

Figure 2(a) shows the results of training this model on the CIFAR 100 dataset[7] for 100 epochs with a batch size of 128. We see that Eve outperforms all other algorithms by a large margin. It quickly surpasses other methods, and achieves a much lower final loss at the end of training.

### 4.2 TRAINING RNNS

We also compare our method with other optimizers for training Recurrent Neural Networks (RNNS). We use the same algorithms as in the previous experiment, and the

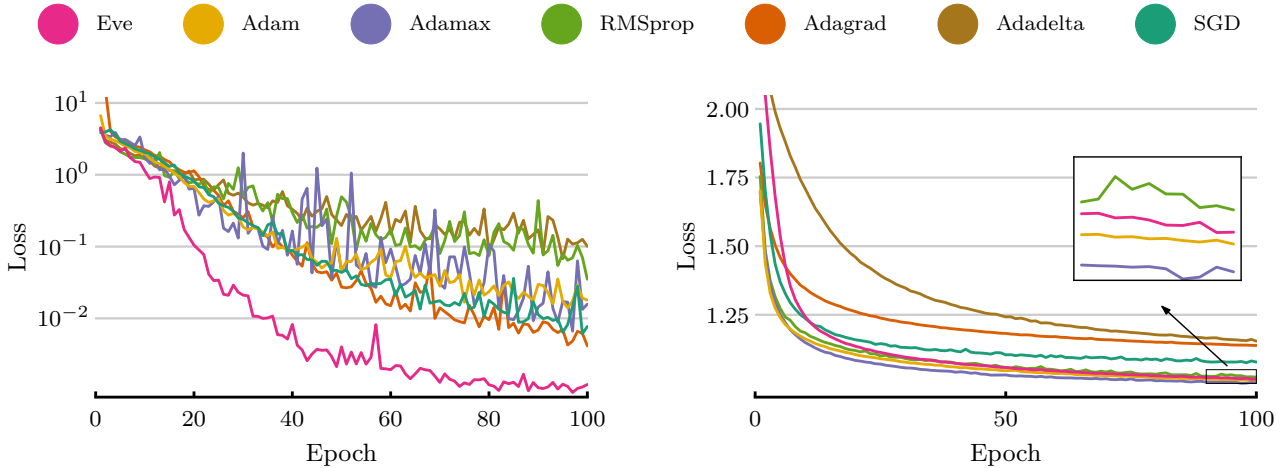

**a.** Residual CNN for image classification.

**b.** RNN for language modeling. Inset is last 10 epochs.

Figure 2: Training loss comparison. In both experiments, Eve achieves similar or lower loss than other optimizers.

learning rate search was conducted over the same set of values.

We construct a RNN for character-level language modeling task on Penn Treebank (PTB)[9]. Specifically, the model consists of a 2-layer character-level Gated Recurrent Unit (GRU)[1] with hidden layers of size 256, with 0.5 dropout between layers. The sequence length is fixed to 100 characters, and the vocabulary is kept at the original size.

The results for training this model are shown in Figure 2(b). Different optimizers performed similarly on this task, with Eve achieving slightly higher loss than Adam and Adamax.

### 4.3 COMPARISON WITH DECAY STRATEGIES

We also empirically compare Eve with three common decay policies: exponential ($\alpha_t = \alpha_1 \exp(-\gamma t)$), $1/t$ ($\alpha_t = \alpha_1/(1+\gamma t)$), and $1/\sqrt{t}$ ($\alpha_t = \alpha_1/\sqrt{1+\gamma t}$). We consider the same CIFAR 100 classification task described in Section 4.1, and use the same CNN model. We applied the three decay policies to Adam, and tuned both the initial learning rate and decay strength. Learning rate was again searched over the same set as in the previous experiments.

For $\gamma$, we searched over a different set of values for each of the decay policies, such that final learning rate after 100 epochs would be $\alpha_1/k$ where $k$ is in $\{1 \times 10^4, 5 \times 10^3, 1 \times 10^3, 5 \times 10^2, 1 \times 10^2, 5 \times 10^1, 1 \times 10^1, 5 \times 10^0\}$.

Figure 3(a) compares Eve with the best exponential decay, the best $1/t$ decay, and the best $1/\sqrt{t}$ decay applied to Adam. We see that using decay closes some of the gap between the two algorithms, but Eve still shows faster convergence. Moreover, using such a decay policy requires careful tuning of the decay strength. As seen in Figure 3(b), for different decay strengths, the performance of Adam can vary a lot. Eve can achieve similar or better performance without tuning an additional hyperparameter.

### 4.4 EFFECT OF HYPERPARAMETERS

In this experiment, we study the behavior of Eve with respect to the two hyperparameters introduced over Adam: $\beta_3$ and $c$. We use the previously presented ResNet model on CIFAR 100, and a RNN model trained for question answering, on question 14 (picked randomly) of the bAbI-10k dataset[16]. The question answering model composes two separate GRUs (with hidden layers of size 256 each) for question sentences, and story passages.

We trained the models using Eve with $\beta_3$ in $\{0, 0.00001, 0.0001, 0.001, 0.01, 0.1, 0.3, 0.5, 0.7, 0.9, 0.99, 0.999, 0.9999, 0.99999\}$, and $c$ in $\{2, 5, 10, 15, 20, 50, 100\}$. For each $(\beta_3, c)$ pair, we picked the best learning rate from the same set of values used in previous experiments. We also used Adam with the best learning rate chosen from the same set as Eve.

Figure 4 shows the loss curves for each hyperparameter pair, and that of Adam. The bold line in the figure is for $(\beta_3, c) = (0.999, 10)$, which are the default values. For these particular cases, we see that for almost all settings of the

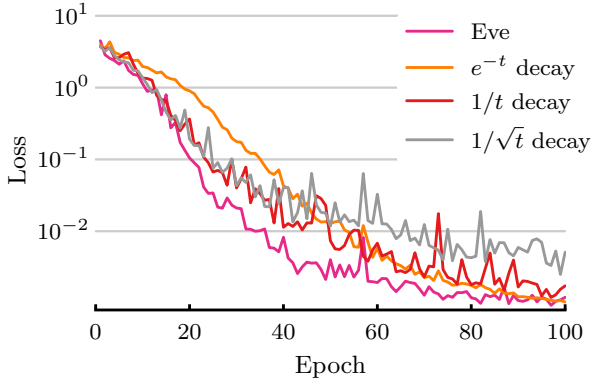
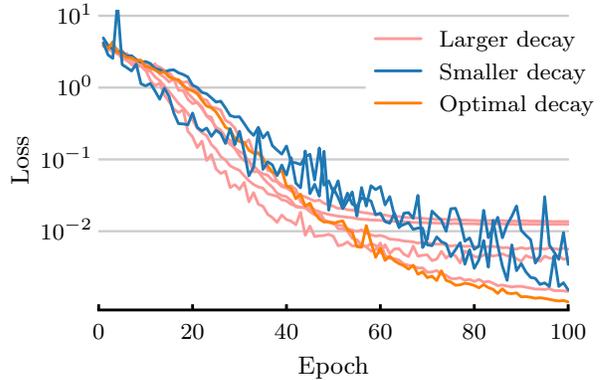

**a.** Eve compared with different decay strategies.

**b.** Adam with different exponential decay strengths.

Figure 3: Results of comparing Eve with learning rate decay strategies. Plot (a) shows the best results for Adam with different decays. The final loss values are similar to that of Eve, but Eve converges faster, and does not require the tuning of an additional parameter. This can be an important factor as shown in plot (b). For suboptimal decay strengths, the performance of Adam varies a lot.

hyperparameters, Eve outperforms Adam, and the default values lead to performance close to the best. In general, for different models and/or tasks, not all hyperparameter settings lead to improved performance over Adam, and we did not observe any consistent trend in the performance across hyperparameters. However, the default values suggested in this paper consistently lead to good performance on a variety of tasks. We also note that the default hyperparameter values were not selected based on this experiment, but through an informal initial search using a smaller model.

## 5 Conclusion and Future Work

We proposed a new algorithm, Eve, for stochastic gradient-based optimization. Our work builds on adaptive methods which maintain a separate learning rate for each parameter, and adaptively tunes the global learning rate using feedback from the objective function. Our algorithm is simple to implement, and is efficient, both computationally, and in terms of memory.

Through experiments with CNNS and RNNS, we showed that Eve outperforms other state of the art optimizers in optimizing large neural network models. We also compared Eve with learning rate decay methods and showed that Eve can achieve similar or better performance with far less tuning. Finally, we studied the hyperparameters of Eve and saw that a range of choices leads to performance improvement over Adam.

One limitation of our method is that it requires knowledge of the global minimum of the objective function. One possible approach to address this issue is to use an estimate of the minimum, and update this estimate as training progresses. This approach has been used when using Polyak step sizes with the subgradient method.

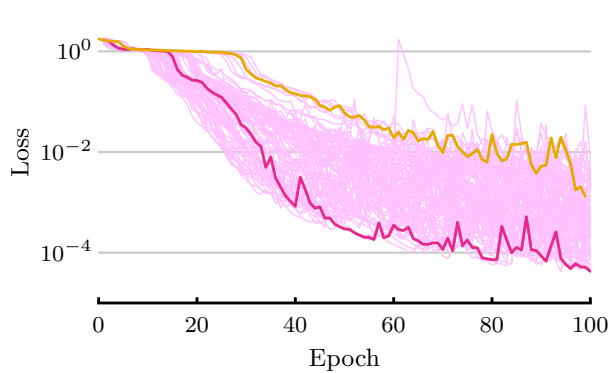 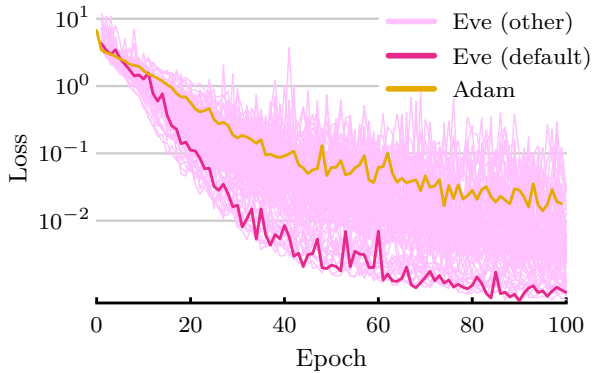

**a.** bAbI-10k, Q-14.

**b.** CIFAR 100.

Figure 4: Loss curves for training with Adam and Eve (with different choices for the hyperparameters).